\documentclass[conference,a4paper]{IEEEtran}
\IEEEoverridecommandlockouts

\usepackage[hidelinks]{hyperref}
\usepackage[cmex10]{amsmath}
\usepackage{amssymb,amsfonts}
\usepackage{dblfloatfix}

\usepackage[ruled,vlined]{algorithm2e}
\usepackage{graphicx}
\graphicspath{{Figures/PDF/}{Figures/PNG/}}

\usepackage{booktabs}
\usepackage{siunitx}
\usepackage[numbers,compress]{natbib}
\usepackage{texnames}
\usepackage{bm,bbm}
\usepackage{orcidlink}

\begin{document}

\title{\uppercase{\ Bridging Spatial and Frequency Views for Disaster Assessment: Benefits and Limitations}
}

\author{	\IEEEauthorblockN{Shikha V.\ Chandel\orcidlink{0009-0002-7592-6009}}
	\IEEEauthorblockA{\textit{College of Science and Technology}\\
        North Carolina A\&T State University\\
        Greensboro, NC\\
		svchandel@aggies.ncat.edu}
	\and
    \IEEEauthorblockN{Yadav Raj \ Ghimire\orcidlink{0009-0002-7592-6009}}
	\IEEEauthorblockA{\textit{College of Science and Technology}\\
        North Carolina A\&T State University\\
		Greensboro, NC\\
		yrghimire@aggies.ncat.edu}
	\and
    \IEEEauthorblockN{Timothy \ Agboada\orcidlink{0009-0007-6505-9088}}
	\IEEEauthorblockA{\textit{Computational Data Science and Engineering}\\
        North Carolina A\&T State University\\
		Greensboro, NC\\
		tagboada@aggies.ncat.edu}
    \and
	\IEEEauthorblockN{Leila Hashemi-Beni\orcidlink{0000-0002-5283-7350}}
	\IEEEauthorblockA{\textit{College of Science and Technology}\\
        North Carolina A\&T State University\\
		Greensboro, NC\\
		lhashemibeni@ncat.edu}
}

\maketitle
\begin{abstract}
Rapid assessment of building damage from satellite imagery is essential for effective disaster response and recovery. While most deep learning methods rely on spatial-domain features, frequency-domain representations can capture complementary structural cues such as debris patterns and collapse-induced textures. This study presents a controlled comparison of spatial-domain, frequency-domain, and dual-domain deep learning approaches for multi-class building damage classification using post-disaster imagery from the xView2 (xBD) dataset. To ensure fairness, all models are built on an EfficientNet-B0 backbone and trained under identical settings, differing only in their input representations and fusion strategies.

Performance is evaluated using accuracy, macro F1-score, per-class metrics, and confusion matrices. Results show that dual-domain models provide measurable improvements over single-domain approaches. The dual spatial configuration achieves the highest test accuracy (0.4688) and lowest loss, while the spatial-only model attains the best macro F1-score (0.4254), indicating more balanced class performance. In contrast, frequency-only models perform worst and exhibit overfitting, suggesting limited generalization.

Despite these gains, all models struggle to detect subtle damage levels, particularly the Minor class, due to class imbalance and fine-grained visual ambiguity. While dual-domain approaches improve detection of severe damage, challenges remain. These findings highlight the benefits and limitations of hybrid representations and motivate future work on data balancing, advanced fusion, and regularization.
\end{abstract}


\thanks{© 2026 IEEE. Published in the Proceedings of the 2026 IEEE International Geoscience and Remote Sensing Symposium (IGARSS 2026), Washington, DC, USA, 9--14 August 2026.

Personal use of this material is permitted. Permission from IEEE must be obtained for all other uses, including reprinting/republishing this material for advertising or promotional purposes, creating collective works for resale or redistribution to servers or lists, or reuse of any copyrighted component of this work in other works.

For permission requests, contact: IEEE Copyrights and Permissions, IEEE Service Center, 445 Hoes Lane, P.O. Box 1331, Piscataway, NJ 08855-1331, USA; Tel: +1 908-562-3966.

This manuscript is the accepted version submitted to arXiv. The final published version will appear in the Proceedings of IGARSS 2026 and will be available through IEEE Xplore. When citing this work, please refer to the published version.}
\section{Introduction}
Recent advances in artificial intelligence (AI), remote sensing (RS), and generative AI (GenAI) have significantly advanced disaster damage assessment by enabling faster, more accurate, and scalable analysis than traditional field-based inspections \cite{blay2025geospatial}. Large-scale benchmark datasets such as xBD have further accelerated progress in automated building damage detection \cite{wang2024deep, gupta2019xbd}. However, most existing approaches rely on spatial domain representations. 

Spatial-domain neural networks primarily learn shape and texture features derived from pixel intensities, which limits their ability to capture subtle structural deformations, debris distributions, and material disruptions caused by disasters \cite{yang2023cdf}. In contrast, frequency-domain representations allow models to explicitly encode high-frequency information associated with edges, cracks, and texture irregularities, which are often strongly correlated with building damage \cite{yang2023cdf}. 

This paper addresses the unexplored benefits and limitations of domain information in disaster damage assessment by conducting a controlled comparison of spatial-only, frequency-only, and dual-domain deep learning architectures for building damage detection. By isolating architectural and representational factors, this study provides systematic insight into the contribution of understanding domain feature extraction in post-disaster remote sensing analysis. 

\section{Related Work}
Rapid and accurate damage assessment is crucial for prioritizing rescue operations and minimizing further losses \cite{khankeshizadeh2025edb}. Automating damage assessment using a CNN (Convolutional Neural Network)-based model on satellite imagery can reduce assessment time compared to traditional methods \cite{benedict2024comparison}.

There are a few publicly available collections of satellite datasets, such as the xBD datasets, which include bounding boxes and descriptions of environmental elements such as smoke, water, lava, and fire \cite{liu2025rescueadi, wang2024deep, gupta2019xbd}. xBD provides pre- and post-event satellite imagery for a variety of disaster events, including building polygons through masks, ordinal damage level labels, and corresponding satellite metadata\cite{wang2024deep, benedict2024comparison, gupta2019xbd}.  

Majorly, prior studies use spatial-domain information, where Convolutional Neural Networks (CNNs) and Vision Transformers (ViTs) extract features from pixel intensities to capture shape, texture, and object geometry \cite{tan2019efficientnet, raghu2021vision, chen2024hrtbda}. U-Net and EfficientNet variants remain popular backbones due to their ability to model both local and global spatial patterns \cite{tan2019efficientnet, wu2021building}. However, spatial-domain models can struggle to capture subtle structural deformations, debris patterns, or fine roof-level damages under complex post-disaster scenarios, which may lead to bias towards no-damage predictions \cite{khankeshizadeh2025edb, wang2024deep}. 

Frequency-domain representations, such as the Fourier and wavelet decompositions, explicitly highlight high-frequency components like edges and texture irregularities, which are often indicative of structural damage \cite{li2023ddformer, chen2024hrtbda}. Hybrid or dual-domain models that combine spatial and frequency features, such as DDFormer and JFSDNet, have been proposed to improve robustness and capture complementary damage cues. However, systematic evaluation in disaster assessment remains limited \cite{li2023ddformer, khankeshizadeh2025edb, zhou2022joint}.

Several recent surveys and comparative studies provide comprehensive evaluations of various deep learning architectures across multiple remote-sensing disaster datasets \cite{wang2024deep, benedict2024comparison}. Furthermore, interdisciplinary methods integrating domain knowledge, crowdsourcing, and 3D machine learning have been shown to enhance both accuracy and interpretability in building damage assessment \cite{kohns2025building}. However, neither study discusses the limitations and benefits of the spatial, frequency, and dual-domain methods in a controlled environment. This leaves a gap in the literature. Hence, this work explicitly evaluates the contribution of frequency information in post-disaster imagery, complementing spatial-domain and hybrid approaches reported in recent literature \cite{khankeshizadeh2025edb, li2023ddformer, chen2024hrtbda}. 

\section{Methodology}
\subsection{Dataset and Preprocessing}

This study is conducted on the xView2 dataset, originally released as part of the DARPA xView2 challenge. The dataset consists of paired pre- and post-disaster satellite images accompanied by annotations describing disaster types and building damage levels \cite{gupta2019xbd}. The data set is organized into four subsets: Train, Tier-3, Hold, and Test. 
\begin{table}[hbt]
	\centering
	\caption{Label distribution for data.}\label{tab:damage_category}
	\begin{tabular}{lccc}
		\toprule
		\textbf{Label} & \textbf{Train} & \textbf{Validation} & \textbf{Test}  \\ 
        \cmidrule(lr){1-1} \cmidrule(lr){2-2} \cmidrule(lr){3-3} \cmidrule(lr){4-4} 
		No Damage & 4742 & 1186 & 873\\
		Destroyed & 1633 & 408 & 620\\
		Minor & 370 & 93 & 121\\
		Major & 589 & 147 & 252\\ \bottomrule
	\end{tabular}
\end{table}

Table~\ref{tab:damage_category}  presents the distribution of the dataset across four damage levels for the training, validation, and test sets. All images are resized to a fixed spatial resolution (256×256) and transformed, ensuring consistency across different network architectures. To isolate the contributions of spatial- and frequency-domain representations, all experiments use only post-disaster RGB imagery, excluding pre-disaster data and additional modalities. The models are trained and validated using the training set and Tier-3 subset, while evaluation is performed on the hold and test subsets. Class imbalance is addressed using Focal Loss with Smoothing. 

\subsection{Architectural Variants}

In this study, three deep learning architectures are evaluated: 

\textbf{1) Spatial-domain model:} This model processes post-disaster RGB images directly in the spatial domain using a convolutional neural network backbone. 


\textbf{2) Frequency-domain model:} This model operates on frequency-domain representations obtained by applying an orthogonally normalized two-dimensional discrete Fourier transform independently to each RGB channel of the post-disaster imagery. The magnitude extraction and optional logarithmic amplitude scaling are used as input to the network. 

        
\textbf{3) Dual-domain model:} This model integrates frequency and spatial-domain information through parallel network branches, respectively. Spatial features extracted from RGB images and frequency features extracted from Fourier magnitude spectra are fused at the feature level before classification. Two dual-domain models are experimented with concatenation, as follows:

1) Frequency Domain + Spatial Domain = Dual Frequency


2) Spatial Domain + Frequency Domain = Dual Spatial


\subsection{Backbone Architecture and Training Protocol}

All architectural variants use the same EfficientNet-B0 convolutional neural network backbone, initialized with no weights, to ensure consistency and a fair comparison. Models are trained using supervised classification with the Adam optimizer, and identical hyperparameters are applied across all experiments. As models are trained for 50 epochs each, the best model is saved based on the F1 macro score on the validation set at each epoch.

\subsection{Evaluation metrics and Analysis}

\begin{figure}[hbt]
	\centering
	\includegraphics[width=.9\linewidth]{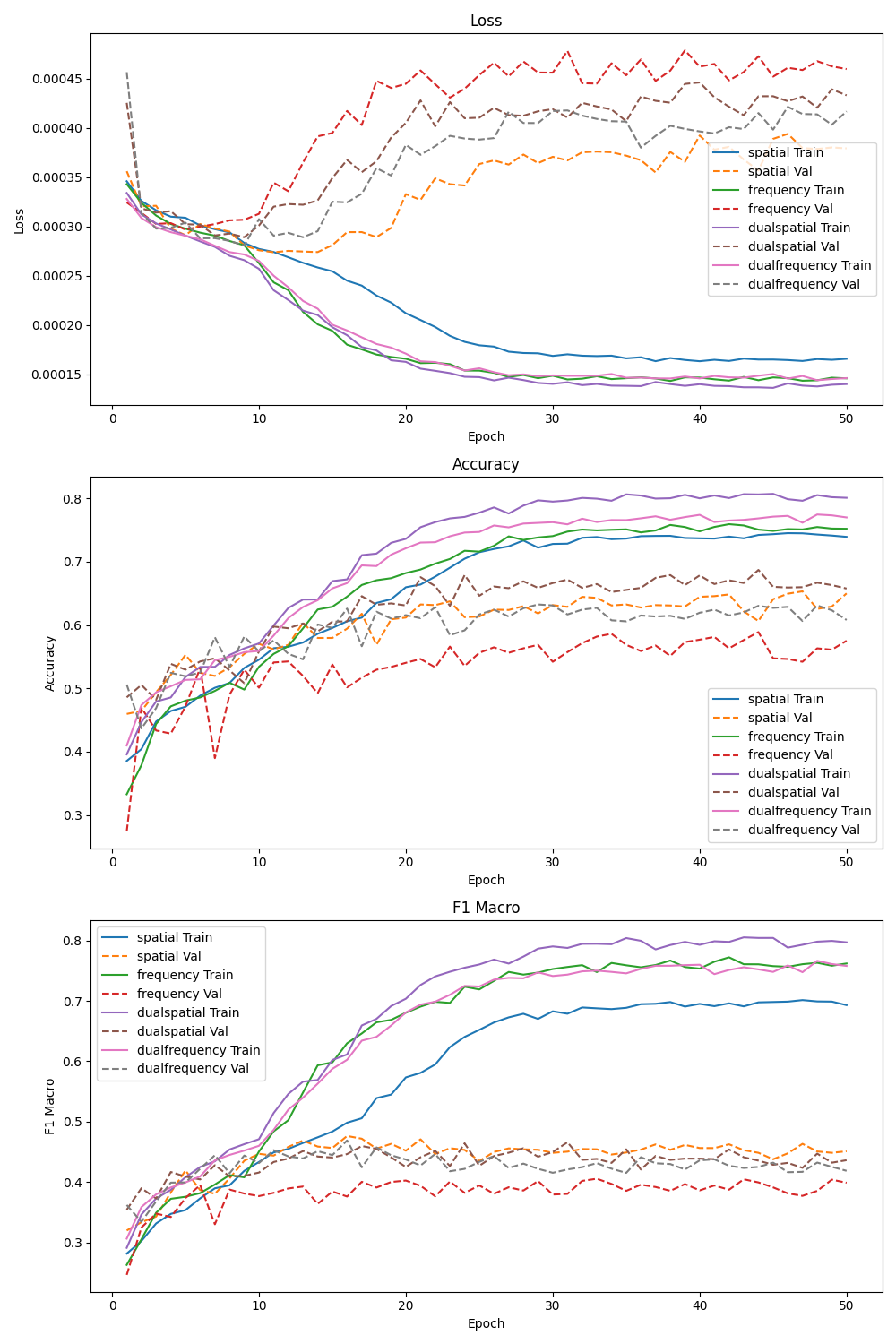}
    \caption{Graphs for Loss, Accuracy, and F1 Macro for each model for 50 epochs for Training and Validation datasets}
	\label{fig:train_comp}
\end{figure}

Training is monitored using "Loss vs. Epochs," "F1 Macro vs. Epochs," and "Accuracy vs. Epochs" graphs as illustrated in Fig.~\ref{fig:train_comp}. Additionally, per-class accuracy during training is used to illustrate the model's learning pattern.
Model performance is evaluated using classification accuracy, macro F1-score, per-class accuracy, and confusion matrices to account for class imbalance in the dataset.

\section{Results}

\subsection {Quantitative Performance Comparison}
The results consistently demonstrate that dual-domain approaches outperform single-domain models in classification performance. In terms of training performance, all models show a steady decrease in loss and a corresponding increase in accuracy and F1-score, indicating successful learning. Among them, the dual-spatial model achieves the highest training accuracy (80\%) and F1-score (0.80), followed by the dual-frequency and frequency models. The spatial-only model exhibits comparatively lower training performance.

However, validation performance provides a more reliable indicator of generalization. The dual-spatial model again achieves the best results, with a validation accuracy of approximately 66\%–68\% and an F1 Score of approximately 0.45–0.47. The spatial-only model performs moderately well, while the dual-frequency model shows slightly lower but competitive performance. In contrast, the frequency-only model consistently underperforms, achieving the lowest validation accuracy (0.55–0.58) and F1-score (0.38–0.40).

These results indicate that integrating spatial and frequency-domain representations enhances classification performance, supporting the hypothesis that dual-domain feature learning is more effective than relying on a single domain.

\subsection {Overfitting and Generalization Analysis}
Despite strong training performance, all models exhibit signs of overfitting, as evidenced by the divergence between training and validation loss after approximately 10–15 epochs. This effect is most pronounced in the frequency-only model, where validation loss increases sharply while training loss continues to decrease.

The spatial-only model also shows moderate overfitting, whereas the dual-domain models demonstrate comparatively better generalization, with a slower increase in validation loss and a smaller gap between training and validation metrics.

Nevertheless, even the best-performing dual-spatial model exhibits a noticeable gap between training and validation F1-scores (0.80 vs. 0.46), suggesting that additional regularization techniques may be necessary to improve generalization.

\subsection{Impact of Domain Representation}
The observed performance differences highlight the importance of feature representation:

{1.} The spatial domain captures structural and contextual information such as shapes and textures, which are critical for identifying building damage.

{2.} The frequency domain captures high-frequency patterns, edges, and periodic structures but may lose spatial localization, leading to weaker standalone performance.

{3.} The dual-domain approach leverages complementary information from both domains, enabling the model to learn richer and more discriminative features.

{4.} The superior performance of the dual-spatial model suggests that effective fusion of spatial and frequency features enhances its ability to capture complex damage patterns that may not be fully represented in a single domain.

\subsection {Test Set Performance Analysis}
\begin{table}[hbt]
	\centering
	\caption{Test Classification Report for Spatial Model}\label{tab:cr_spatial}
	\begin{tabular}{l S[table-format=2.2] S[table-format=2.2] S[table-format=2.2] S[table-format=2.2]}
		\toprule
		\textbf{Spatial} & \textbf{Precision} & \textbf{Recall} & \textbf{Macro-F1} \\ \cmidrule(lr){2-2}  \cmidrule(lr){3-3} \cmidrule(lr){4-4}\\
		No Damage & 0.85  & 0.33  & 0.48  \\
		Minor & 0.13 & 0.50  & 0.21 \\
		Major & 0.43  & 0.60  & 0.50 \\
		Destroyed & 0.45  & 0.59  & 0.51 \\ 
        \bottomrule
	\end{tabular}
\end{table}

\begin{table}[hbt]
	\centering
	\caption{Test Classification Report for Frequency Model}\label{tab:cr_frequency}
	\begin{tabular}{l S[table-format=2.2] S[table-format=2.2] S[table-format=2.2] S[table-format=2.2]}
		\toprule
		\textbf{frequency} & \textbf{Precision} & \textbf{Recall} & \textbf{Macro-F1} \\ \cmidrule(lr){2-2}  \cmidrule(lr){3-3} \cmidrule(lr){4-4}\\
		No Damage & 0.64  & 0.55  & 0.59  \\
		Minor & 0.00 & 0.00  & 0.00 \\
		Major & 0.23  & 0.30  & 0.26 \\
		Destroyed & 0.33  & 0.41  & 0.37 \\ 
        \bottomrule
	\end{tabular}
\end{table}

\begin{table}[hbt]
	\centering
	\caption{Test Classification Report for Dual Spatial-Frequency Domain Model}\label{tab:cr_spatial_dual}
	\begin{tabular}{l S[table-format=2.2] S[table-format=2.2] S[table-format=2.2] S[table-format=2.2]}
		\toprule
		\textbf{} & \textbf{Precision} & \textbf{Recall} & \textbf{Macro-F1} \\  \cmidrule(lr){2-2}  \cmidrule(lr){3-3} \cmidrule(lr){4-4}\\
		No Damage & 0.73  & 0.48  & 0.58  \\
		Minor & 0.00 & 0.00  & 0.00 \\
		Major & 0.18  & 0.20  & 0.19 \\
		Destroyed & 0.40  & 0.71  & 0.51 \\ 
        \bottomrule
	\end{tabular}
\end{table}

\begin{table}[hbt]
	\centering
	\caption{Test Classification Report for Dual Frequency-Spatial Domain Model}\label{tab:cr_frequency_dual}
	\begin{tabular}{l S[table-format=2.2] S[table-format=2.2] S[table-format=2.2] S[table-format=2.2]}
		\toprule
		\textbf{} & \textbf{Precision} & \textbf{Recall} & \textbf{Macro-F1} \\  \cmidrule(lr){2-2}  \cmidrule(lr){3-3} \cmidrule(lr){4-4}\\
		No Damage & 0.65  & 0.52  & 0.58  \\
		Minor & 0.17 & 0.25  & 0.20 \\
		Major & 0.25  & 0.30  & 0.27 \\
		Destroyed & 0.35  & 0.41  & 0.38 \\ 
        \bottomrule
	\end{tabular}
\end{table}

\begin{table}[hbt]
	\centering
	\caption{Loss, Accuracy and Macro-F1 across Domains for Test Dataset}\label{tab:test_acc}
	\begin{tabular}{l S[table-format=2.2] S[table-format=2.2] S[table-format=2.2]}
		\toprule
		\textbf{Models} & \textbf{Loss} & \textbf{Accuracy} & \textbf{Macro-F1} \\ \cmidrule(lr){1-1} \cmidrule(lr){2-2}  \cmidrule(lr){3-3} \cmidrule(lr){4-4}\\
		Spatial &  0.0410 & 0.4531  & 0.4254  \\
		  Frequency & 0.0420 & 0.4375  & 0.3049 \\
		Dual Spatial & 0.0351  & 0.4688  & 0.3207 \\
        Dual Frequency & 0.0380  & 0.4375  & 0.3568 \\
        \bottomrule
	\end{tabular}
\end{table}

The performance of the evaluated models on the test dataset reveals notable differences in their ability to generalize across damage categories. As shown in Table \ref{tab:test_acc}, the Dual Spatial model achieves the highest overall accuracy (0.4688) and lowest loss (0.0351), indicating improved predictive capability compared to other models. However, the Spatial model attains the highest macro-F1 score (0.4254), suggesting better class-wise balance despite slightly lower accuracy. In contrast, the Frequency model exhibits the weakest performance, with the lowest macro-F1 score (0.3049), highlighting its limited effectiveness when used independently.

A closer examination of class-wise metrics provides further insights. The Spatial model performs well on structurally prominent classes, achieving F1-scores of 0.50 and 0.51 for Major and Destroyed categories, respectively. However, it demonstrates poor recall (0.33) for the No Damage class despite high precision (0.85), indicating a tendency to miss many true negatives. Additionally, the model struggles with the Minor class (F1: 0.21), reflecting difficulty in identifying subtle damage patterns.

The Frequency model shows consistently weak performance across all classes, including a complete failure to classify the Minor category (F1: 0.00). This suggests that frequency-domain features alone lack sufficient discriminative power for capturing complex structural variations in building damage.

The Dual Spatial model improves detection of severe damage, achieving the highest recall (0.71) for the Destroyed class and maintaining a strong F1-score (0.51). It also performs moderately well for the No Damage class (F1: 0.58). However, similar to the frequency model, it fails to capture the minor class and performs poorly on the Major category (F1: 0.19), indicating persistent challenges in intermediate damage classification.

In comparison, the Dual Frequency model demonstrates more balanced performance across all classes, achieving moderate F1 scores for the Minor (0.20), Major (0.27), and Destroyed (0.38). Although it does not achieve the highest performance in any single category, its ability to handle all classes more consistently suggests improved generalization.

Overall, the results indicate that while dual-domain approaches enhance performance in specific categories, particularly for severe damage, they do not fully address class imbalance, especially for the Minor class. The findings also reinforce that spatial features are critical for capturing structural damage, while frequency features provide complementary information but are insufficient when used in isolation. These results partially support the hypothesis, demonstrating that dual-domain models improve certain aspects of classification performance, but further refinement is required to achieve robust, balanced predictions across all damage levels.

\section{Discussion}

The experimental results across training, validation, and test datasets collectively demonstrate that dual-domain learning provides measurable benefits over single-domain approaches, while also revealing important limitations. From the learning curves, dual-domain models, particularly the dual spatial configuration, consistently achieve higher training accuracy and F1-scores and exhibit better generalization than frequency-only models. This is further supported by the test set results, where the Dual Spatial model achieves the highest accuracy (0.4688) and lowest loss (0.0351), indicating stronger predictive capability. However, the Spatial model attains the highest macro-F1 score (0.4254), suggesting that it maintains better balance across classes. In contrast, the frequency-only model performs the worst across all metrics and shows clear signs of overfitting and poor generalization, confirming that frequency-domain features alone are insufficient for robust damage classification. These findings support the hypothesis that combining spatial and frequency representations enhances performance, particularly in capturing complex and severe damage patterns.

Despite these improvements, the results also highlight persistent challenges, most notably class imbalance and difficulty in detecting subtle damage levels. All models struggle significantly with the Minor class, with near-zero or very low F1-scores in several cases, indicating that neither spatial nor frequency features individually or combined are sufficient to capture fine-grained damage distinctions reliably. While the Dual Spatial model excels in identifying severe damage (e.g., high recall for the Destroyed class), it does so at the expense of intermediate classes. In contrast, the Dual Frequency model offers more balanced but lower overall performance. Additionally, the gap between training and validation metrics suggests that overfitting persists, even for dual-domain approaches. Therefore, although dual-domain models improve classification performance, further enhancements such as data balancing, advanced feature fusion strategies, and regularization techniques are necessary to achieve robust, real-world applicability in post-disaster damage assessment systems.

\section{Conclusion and Future Work}
This study presents a controlled empirical evaluation of spatial, frequency, and dual-domain representations for building damage classification, using a consistent CNN backbone. It demonstrates that dual-domain learning improves overall predictive performance, with the dual spatial–frequency model achieving the highest test accuracy and highlighting the benefit of integrating complementary feature spaces. Nevertheless, the spatial-domain model remains highly competitive, achieving a higher macro-F1 score and more stable class-wise performance, indicating that spatial features alone provide robust and reliable representations for structural damage assessment. By contrast, the frequency-domain model consistently underperforms on unseen data due to limited generalization and susceptibility to overfitting. Although dual-domain fusion enhances performance in specific cases, particularly for severe damage, it does not uniformly improve results across all classes, suggesting that simple fusion strategies may not fully leverage the complementary nature of spatial and frequency information. A major challenge across all models is poor performance on minority and intermediate damage classes, especially the Minor and Destroyed categories, which significantly affects macro-F1 scores. It underscores the need for class-sensitive evaluation metrics beyond accuracy. To address these limitations, future work should focus on improving class balance and representation learning through targeted data augmentation, adaptive or attention-based fusion mechanisms, and advanced architectures such as transformer-based or hybrid models, along with class-aware loss functions or reweighting strategies to enhance performance on imbalanced data, ultimately improving the robustness and real-world applicability of automated disaster damage assessment systems.

\section{Acknowledgement}
This work is supported by NSF AWARD 2401942 and NASA AWARD 80NSSC23M0051.

\small
\bibliographystyle{IEEEtranN}
\bibliography{references}

@article{kohns2025building,
  title={Building damage assessment in natural disasters: A trans-and interdisciplinary approach combining domain knowledge, 3D machine learning, and crowdsourcing},
  author={Kohns, Julia and Zahs, Vivien and Klonner, Carolin and H{\"o}fle, Bernhard and Stempniewski, Lothar and Stark, Alexander},
  journal={Progress in Disaster Science},
  pages={100427},
  year={2025},
  publisher={Elsevier}
}

@article{liu2025rescueadi,
  title={RescueADI: adaptive disaster interpretation in remote sensing images with autonomous agents},
  author={Liu, Zhuoran and Zhao, Danpei and Yuan, Bo and Jiang, Zhiguo},
  journal={IEEE Transactions on Geoscience and Remote Sensing},
  year={2025},
  publisher={IEEE}
}

@article{khankeshizadeh2025edb,
  title={EDB-HSTEU-Net: Earthquake-Damaged Building Detection using A Novel Hybrid Swin Transformer Efficient U-Net (HSTEU-Net) and Transfer Learning Techniques from Post-event VHR Remote Sensing Data},
  author={Khankeshizadeh, Ehsan and Mohammadzadeh, Ali and Jamali, Sadegh},
  journal={Journal of Building Engineering},
  pages={112889},
  year={2025},
  publisher={Elsevier}
}

@article{wang2024deep,
  title={Deep learning models for hazard-damaged building detection using remote sensing datasets: A comprehensive review},
  author={Wang, Lili and Wu, Jidong and Yang, Youtian and Tang, Rumei and Ya, Ru},
  journal={IEEE Journal of Selected Topics in Applied Earth Observations and Remote Sensing},
  year={2024},
  publisher={IEEE}
}

@article{li2023ddformer,
  title={DDFormer: A dual-domain transformer for building damage detection using high-resolution SAR imagery},
  author={Li, Tianyang and Wang, Chao and Zhang, Hong and Wu, Fan and Zheng, Xiaohan},
  journal={IEEE Geoscience and Remote Sensing Letters},
  volume={20},
  pages={1--5},
  year={2023},
  publisher={IEEE}
}

@article{chen2024hrtbda,
  title={HRTBDA: a network for post-disaster building damage assessment based on remote sensing images},
  author={Chen, Fang and Sun, Yao and Wang, Lei and Wang, Ning and Zhao, Huichen and Yu, Bo},
  journal={International Journal of Digital Earth},
  volume={17},
  number={1},
  pages={2418880},
  year={2024},
  publisher={Taylor \& Francis}
}

@inproceedings{benedict2024comparison,
  title={Comparison on Difference Deep Learning Models for Building Damage Assessment using xBD Dataset},
  author={Benedict, Reiko and Winartio, Reinhart Brilian and Adinata, Muhammad Faisal and Irwansyah, Edy and others},
  booktitle={2024 Arab ICT Conference (AICTC)},
  pages={181--186},
  year={2024},
  organization={IEEE}
}

@inproceedings{tan2019efficientnet,
  title={Efficientnet: Rethinking model scaling for convolutional neural networks},
  author={Tan, Mingxing and Le, Quoc},
  booktitle={International conference on machine learning},
  pages={6105--6114},
  year={2019},
  organization={PMLR}
}

@article{gupta2019xbd,
  title={xbd: A dataset for assessing building damage from satellite imagery},
  author={Gupta, Ritwik and Hosfelt, Richard and Sajeev, Sandra and Patel, Nirav and Goodman, Bryce and Doshi, Jigar and Heim, Eric and Choset, Howie and Gaston, Matthew},
  journal={arXiv preprint arXiv:1911.09296},
  year={2019}
}

@article{wu2021building,
  title={Building damage detection using U-Net with attention mechanism from pre-and post-disaster remote sensing datasets},
  author={Wu, Chuyi and Zhang, Feng and Xia, Junshi and Xu, Yichen and Li, Guoqing and Xie, Jibo and Du, Zhenhong and Liu, Renyi},
  journal={Remote Sensing},
  volume={13},
  number={5},
  pages={905},
  year={2021},
  publisher={MDPI}
}

@article{raghu2021vision,
  title={Do vision transformers see like convolutional neural networks?},
  author={Raghu, Maithra and Unterthiner, Thomas and Kornblith, Simon and Zhang, Chiyuan and Dosovitskiy, Alexey},
  journal={Advances in neural information processing systems},
  volume={34},
  pages={12116--12128},
  year={2021}
}

@article{yang2023cdf,
  title={CDF-net: A convolutional neural network fusing frequency domain and spatial domain features},
  author={Yang, Aitao and Li, Min and Wu, Zhaoqing and He, Yujie and Qiu, Xiaohua and Song, Yu and Du, Weidong and Gou, Yao},
  journal={IET computer vision},
  volume={17},
  number={3},
  pages={319--329},
  year={2023},
  publisher={Wiley Online Library}
}

@article{zhou2022joint,
  title={Joint frequency-spatial domain network for remote sensing optical image change detection},
  author={Zhou, Yuan and Feng, Yanjie and Huo, Shuwei and Li, Xiaofeng},
  journal={IEEE Transactions on Geoscience and Remote Sensing},
  volume={60},
  pages={1--14},
  year={2022},
  publisher={IEEE}
}

@article{blay2025geospatial,
  title={Geospatial and Deep Learning Approaches for Modeling Floodwater Depth in Urbanized Areas},
  author={Blay, Jeffrey and Hashemi-Beni, Leila},
  journal={Remote Sensing},
  volume={18},
  number={1},
  pages={60},
  year={2025},
  publisher={MDPI}
}

\end{document}